\pdfoutput=1

\documentclass[11pt]{article}

\usepackage[final]{acl}

\usepackage{times}
\usepackage{latexsym}
\usepackage{amsmath}
\usepackage{graphicx}
\usepackage{xcolor}
\usepackage{amssymb}
\usepackage[T1]{fontenc}

\usepackage[utf8]{inputenc}

\usepackage{microtype}

\usepackage{inconsolata}

\title{LeCoPCR: Legal Concept-guided Prior Case Retrieval for \\ European Court of Human Rights cases}

\author{Santosh T.Y.S.S, Isaac Misael Olguín Nolasco, Matthias Grabmair \\ School of Computation, Information, and Technology; \\
Technical University of Munich, Germany \\ \ }

\begin{document}
\maketitle
\begin{abstract}
Prior case retrieval (PCR) is crucial for legal practitioners to find relevant precedent cases given the facts of a query case. Existing approaches often overlook the underlying semantic intent in determining relevance with respect to the query case. In this work, we propose LeCoPCR, a novel approach that explicitly generate intents in the form of legal concepts from a given query case facts and then augments the query with these concepts to enhance models understanding of semantic intent that dictates relavance. To overcome the unavailability of annotated legal concepts, we employ a weak supervision approach to extract key legal concepts from the reasoning section using Determinantal Point Process (DPP) to balance quality and diversity. Experimental results on the ECtHR-PCR dataset demonstrate the effectiveness of leveraging legal concepts and DPP-based key concept extraction.
\end{abstract}

\section{Introduction}
Prior Case Retrieval (PCR) involves finding relevant precedent cases for an undecided case query, assisting legal practitioners in citing pertinent precedents to establish applicable laws and craft their arguments \cite{turtle1995text}. The exponential growth of legal cases has made navigating the corpus for relevant prior cases increasingly challenging, attracting significant attention from both the legal and IR communities \cite{locke2022case}. Compared to traditional text retrieval, PCR poses unique challenges due to longer, more complex queries, and the abstract nature of relevance in legal tasks \cite{van2017concept}.

Several PCR datasets have been constructed recently, such as COLIEE \cite{kano2019coliee}, IRLeD \cite{mandal2017overview}, IL-PCR \cite{joshi2023u} in English from Canada and India, CAIL-SCM \cite{xiao2019cail2019} and LeCaRD \cite{ma2021lecard} from China in Chinese. However, \citet{santosh2024pcr} pointed out that the existing PCR datasets from common law systems do not simulate a realistic setting, as their queries use complete case documents while only masking references to prior cases, resulting in queries containing legal reasoning which is not yet available for an undecided case. To address this, they curated a new PCR dataset based on European Court of Human Rights (ECtHR) jurisdiction, which explicitly separate facts from reasoning and uses only facts of the case as query, treating cited cases in the reasoning section of the query case as relevant precedents.

Earlier methods for PCR relied on expert knowledge based methods with ontologies \cite{saravanan2009improving}, thesaurus \cite{ca2006thesaurus}, legal issue decomposition \cite{zeng2005knowledge}, citation networks \cite{lopez2013link,minocha2015finding}. Recent years have seen the emergence of deep learning approaches with text representations \cite{shao2020bert,rossi2019legal}, domain-aware pre-training \cite{xiao2021lawformer}, incorporating structural information \cite{li2023sailer,ma2023incorporating}, summarization to account for longer length \cite{askari2021combining,tran2019building,zhou2023boosting}, hybrid methods combining text and citation networks \cite{bhattacharya2022legal}. A key challenge in PCR lies in determining the relevance of a prior case, which typically is to substantiate arguments based on shared legal principles or norms and to strengthen the reasoning in the query case. Capturing this semantic intent—the conceptual relationship between the prior and query cases—is essential for accurate retrieval. However, existing methods often rely on the assumption that models can implicitly infer this intent from data, leading to suboptimal performance.

Identifying relevant prior cases to be cited in the query case requires forward-looking anticipation. Users identify the legal principles central to their arguments and search for prior cases that reinforce them. The most authoritative cases, which discuss or clarify the required legal concepts, are then selected for citation. While anticipatory retrieval aims to maximize recall by identifying all potentially relevant cases, determining authoritativeness ensures precision by focusing on the most impactful cases.

In this work, we focus on improving recall in prior case retrieval by explicitly capturing the semantic intent based on specific legal concepts or norms that dictate the relevance of a case to the query. Since the reasoning which provides this semantic intent is not explicitly available at inference time for the query case, unlike prior methods that implicitly expect the model to learn this intent, we make the model deduce the reasoning required for the query case and leverage that reasoning to identify relevant prior cases. However, generating complete reasoning poses two challenges: (i) the lengthy reasoning sections in legal cases exceed the length limitations of current seq2seq models, and (ii) much of the detailed argumentation is irrelevant to the task of citing a specific case, acting as noise. To address this, we hypothesize that condensing the reasoning into key concepts provides a stronger signal for relevance. Therefore, we focus on training a seq2seq model to generate these key concepts from the query case facts as input.

To address the lack of annotated data for legal concepts, we adopt a weak supervision approach to extract key concepts from the reasoning section. We use Determinantal Point Process (DPP) \cite{borodin2009determinantal} to select key concepts from a list of candidates, balancing  quality and diversity. DPP has proven effective in various diversity promoting tasks, such as recommendation \cite{wilhelm2018practical}, answer retrieval \cite{nandigam2022diverse}, text summarization \cite{cho2019improving,cho2019multi,li2019conclusion}, video summarization \cite{zhang2016video,sharghi2017query}. These concepts are then used to train a seq2seq model to generate them from given query case facts during inference. We then augment the query with these generated concepts  to retrieve similar precedent cases, akin to query expansion method \cite{rocchio1971relevance,lavrenko2001relevance,liu2022query}. Our experiments on the ECtHR-PCR dataset demonstrate the effectiveness of explicitly leveraging legal concepts to enhance the model's understanding of relevance. Further, DPP-based concept extraction outperforms other approaches in retrieval performance validated through oracle ablations, attributed to the selection of representative diverse concepts.

\section{Our Method: LeCoPCR}
LeCoPCR initially generates legal concepts from the factual description of a query case and then uses these concepts along with the query to retrieve relevant prior cases. LeCoPCR enhances retrieval by deducing necessary legal concepts required to to establish precedence and provide argumentation for the query case. These legal concepts serve as the intents behind citing a specific prior case in the reasoning section of a query case.

\subsection{Legal Concept Generation}
We use a seq2seq model to generate legal concepts from case facts descriptions. Due to the lack of annotated datasets with legal concepts, we adopt a weak supervision method, wherein we extract key concepts from the reasoning section of cases as silver data for training. It's worth noting that the query comprises only factual descriptions, excluding the reasoning section, which is unavailable prior to the verdict. To extract key legal concepts from the reasoning section, we first identify candidate concepts expressed as noun phrases using part-of-speech tagging and noun phrase chunking with NLTK package \cite{bird2009natural} using the regular expression \texttt{<NN.*|JJ>*<NN.*>} from \citet{bennani2018simple}. Subsequently, we use determinantal point process (DPP) \cite{borodin2009determinantal} to select a subset from these candidate concepts, ensuring a balance between relevance and diversity.

\noindent \emph{DPP for key concept selection:} Let $S = {1,\ldots, |S|}$ denote a finite set of extracted candidate concepts. Formally, DPP defines a probability distribution over an exponential number of sets (all $2^{|S|}$ subsets of $S$), parameterized by a single $|S|\times|S|$ positive semi-definite kernel matrix, denoted as $L$. If $k$ is a random set of elements drawn from $S$, then the probability of selecting that subset is given by determinants of sub-matrices of $L$:
\begin{equation*}
\resizebox{\linewidth}{!}{$
p(k ; L) = \frac{\text{det}(L_k)}{\text{det}(L+I)}  \quad \because \sum_{k}\text{det}(L_k) = \text{det}(L + I)
$}
\end{equation*}
Here, $\text{det}(.)$ is the determinant of a matrix, $I$ is the identity matrix and $L_k$ is a sub-matrix of $L$ containing only entries indexed by elements of $k \subseteq S$. \citet{kulesza2012determinantal} provide a decomposition of the $L$-ensemble matrix as a gram matrix, allowing the modeling of relevance and dissimilarity independently and combining them into a single unified formulation with $L_{ij} = q_i\!\cdot s_{ij}\!\cdot q_j$, where $q_i \in \mathbb{R}^+$ is a positive real number indicating the quality/relevance of concept candidate $i$, and $s_{ij}$ captures the similarity between concepts $i$ and $j$.

To understand why $\text{det}(L_k)$ serves as a balanced measure of quality and diversity for a selected set, consider a subset $Y = \{i, j\}$ of elements. The probability of choosing this subset is given as: 
\begin{equation*}
\resizebox{0.7\linewidth}{!}{$
\begin{aligned}
    P(Y ; L) &\propto \det(L_Y) \\
    &= \begin{bmatrix}
    q_i \cdot s_{ii} \cdot q_i & q_j \cdot s_{ij} \cdot q_j \\
    q_j \cdot s_{ji} \cdot q_i & q_j \cdot s_{jj} \cdot q_j
    \end{bmatrix} \\
    &= q_i^2 \cdot q_j^2 \cdot (1- s_{ij}^2)
\end{aligned}
$}
\end{equation*}
If candidate elements are highly relevant, any subset containing them will have a high probability. Conversely, if two candidate elements are similar, any set containing both will have a low probability. Geometrically, this can be interpreted as the squared volume of the space spanned by candidate concept vectors of $Y$, where quality indicates vector length and similarity represents the angle between vectors. This determinant expression turns more complex for larger matrices but a similar intuition holds there. In our case, considering each element as a candidate concept, the final subset achieving the highest probability will include a set of highly relevant concepts while maintaining diversity among them via pairwise repulsion.

\noindent \emph{Deriving similarity and relevance values:} We use LegalBERT \cite{chalkidis2020legal} to obtain the [CLS] representation of each concept and compute their similarities using cosine similarity. Inspired by \citet{zhang2022mderank}, we avoid computing relevance as cosine similarity directly between a candidate concept and its source paragraph due to potential sequence length mismatch. Instead, we adopt a paragraph-paragraph relevance computation approach. Here, we mask the candidate concept from its source paragraph and compute the similarity score between the masked paragraph and the normal paragraph using LegalBERT and take the complement of this score as the relevance score. This approach assumes that the semantic meaning of the masked document remains largely unchanged if only a trivial concept is masked. Additionally, we consider the position of concepts relative to the citation marker in the reasoning section as indicative of their importance, potentially signaling derivation from the cited case. To incorporate this information, we introduce position regularization $\rho_i$ for each candidate concept, inspired by \citet{florescu2017position} which augments the relevance score by multiplying with $\rho_i$:
\begin{equation*}
  r_i = \hat r_i.\rho_i  \quad \text{where} \quad \rho_i = \text{softmax}(e^{1/k})
\end{equation*}
where $k$ indicates the distance between the $i$-th candidate concept and the nearest citation marker.

\noindent \emph{DPP Inference} The MAP inference for DPP involves sub-modular maximization, which is NP-hard \cite{ko1995exact}. Therefore we use greedy algorithm for faster inference \cite{chen2018fast}. This algorithm begins with an empty set and iteratively adds one concept to the selected set. The chosen concept $c$ in each iteration is the one that maximizes the determinant value when added to the current selected set.
\begin{equation*}
\resizebox{0.7\linewidth}{!}{$
\begin{aligned}
    c &= \arg\max_{i \in S-Y } \left[ f(Y \cup \{i\}) - f(Y) \right] \\
    & \text{where} \quad f(Y) = \log \det(L_Y) 
\end{aligned}
$}
\end{equation*}

\subsection{Retrieval Model} Once legal concepts are generated for a given query case, we concatenate them with the facts to create an effective query input for the retriever. LeCoPCR, being model-agnostic, is demonstrated using both the lexical-based BM25 \cite{robertson2004simple} and neural-based bi-encoder \cite{karpukhin2020dense} as retrievers. In the bi-encoder approach, we transform the query and candidate cases into dense representations independently and compute relevance scores using the dot product between them. We train the dense model using contrastive loss, which pulls together the query and relevant cases while pushing away irrelevant ones. Irrelevant cases are  sampled randomly from the candidate pool which are not relevant to the query. During training, we utilize extracted concepts for query augmentation. However, during inference, we use generated legal concepts, which may exhibit lower quality compared to the extracted ones, leading to exposure bias in the retriever. To mitigate this issue, we employ a hybrid training setup where we augment queries with noisy concepts sampled from other documents, along with the golden extracted concepts to make model robust to noisy generations during inference.  


\section{Experiments}
We use the ECtHR-PCR dataset \cite{santosh2024pcr}, consisting of 15,204 cases, chronologically split into training (9.7k, 1960–2014), development (2.1k, 2015–2017), and test (3.2k, 2018–2022) sets. All the cases preceding the date of query case serve as candidate cases, resulting in an average of 14,101.2 candidates per test query. On average, there are 12 relevant cases per test query. We report Recall@k and Mean Average Precision (MAP). Recall@k measures the proportion of relevant documents ranked in the top-k candidates, with k values of 50, 100, 500 and 1000. We report the average Recall@k across all instances. MAP calculates the mean of the Average Precision scores for each instance, where Average Precision is the average of Precision@k scores for every rank position of each relevant document. Higher scores indicate better performance. We employ LongT5 \cite{guo2022longt5} for legal concept generation and Longformer \cite{beltagy2020longformer} for dense retriever, accounting for longer case documents. Detailed implementation details are provided in App. \ref{impl}.

\subsection{Results}
From Table \ref{results-main}, it's evident that LeCoPCR enhances performance for both BM25 and Longformer, suggesting that legal concepts serve as a bridge to learn relevance signals. Additionally, we present oracle values, which use the extracted concepts from the reasoning section of the query case, forming the upper bound. While the oracle concepts exhibit significant potential for improvement, the modest enhancement in LeCoPCR can be attributed to the inclusion of noisy concepts generated during inference, which may provide incorrect relevance signals. Training with noisy samples under the hybrid setup (HT) with Longformer yields better performance than  LeCoPCR, indicating that exposure to concepts of varying quality enhances the model's robustness.
These results also indicate scope for further improvement in concept generation models. 

\begin{table}[!ht]
\scalebox{0.8}{
\begin{tabular}{|l|c|c|c|c|r|}
\hline
\textbf{}    & \textbf{R50} & \textbf{R100} & \textbf{R500} & \textbf{R1K} & \multicolumn{1}{l|}{\textbf{MAP}} \\ \hline
BM25         & 21.84        & 27.49         & 48.62         & 60.88        & 9.54                              \\ 
+ LeCoPCR & 22.52        & 28.42         &  49.86         & 62.22        & 9.78                             \\ 
\textcolor{gray}{+ Oracle}    & \textcolor{gray}{31.26}        & \textcolor{gray}{38.39}         & \textcolor{gray}{60.79}         & \textcolor{gray}{69.05}        & \textcolor{gray}{14.19}                             \\ \hline
Longformer          & 23.98        & 33.97         & 63.41         & 76.10        & 11.59                             \\ 
+ LeCoPCR & 25.22        & 35.16         & 64.86         & 78.23        & 12.46                             \\ 
+ LeCoPCR-HT & 26.47        & 38.62         & 67.14         & 79.39        & 13.68                             \\ 

\textcolor{gray}{+ Oracle}     & \textcolor{gray}{32.89}        & \textcolor{gray}{44.20}          & \textcolor{gray}{72.68}         & \textcolor{gray}{82.66}        & 
\textcolor{gray}{16.22}                             \\ \hline
\end{tabular}}
\caption{Performance on ECHR-PCR dataset. RK, HT indicate Recall@k and Hybrid training respectively.}
\label{results-main}
\end{table}

\noindent \textbf{Ablation on Concept Extraction:} We compare our DPP-based method to different legal concept extraction techniques such as word-based TF-IDF and phrase-based methods like TextRank \cite{mihalcea2004textrank}, KeyBERT \cite{grootendorst2020keybert}, and MDERank \cite{zhang2022mderank}. We ablate DPP method by removing position regularization and MDE rank-based relevance computation, computing it based on cosine similarity between concept and document. Performance is reported using BM25, where the query is augmented with these extracted oracle concepts in Table \ref{ablation-ext}. Using concepts helps identify relevance signals better, as these legal concepts serve as an intent to cite relevant prior cases. TextRank outperforms TF-IDF, indicating that span-based legal concepts provide better signals. KeyBERT performs better than TextRank, highlighting the importance of semantic relatedness in extracting better concepts. MDERank surpasses KeyBERT, suggesting that document-to-document matching is more effective. Even in our method, with the removal of MDE, performance decreases. Removing position regularization also reduces performance, supporting our hypothesis that concepts closer to citation markers are crucial.

\begin{table}[!ht]
\centering
\scalebox{0.82}{
\begin{tabular}{|l|c|c|c|c|c|}
\hline
\textbf{}        & \multicolumn{1}{l|}{\textbf{R50}} & \multicolumn{1}{l|}{\textbf{R100}} & \multicolumn{1}{l|}{\textbf{R500}} & \multicolumn{1}{l|}{\textbf{R1K}} & \multicolumn{1}{l|}{\textbf{MAP}} \\ \hline
-         & 21.84        & 27.49         & 48.62         & 60.88        & 9.54                              \\ 
TF-IDF           & 22.93                             & 28.50                               & 50.59                              & 61.94                             & 11.04                             \\ 
TextRank  & 23.46 &  29.83 & 51.26 &  62.24  & 11.36 \\
KeyBERT          & 24.25                             & 31.43                              & 52.36                              & 63.18                             & 11.52                             \\ 
MDERank          & 26.05                             & 32.34                              & 53.11                              & 64.75                             & 11.69                             \\  \hline
DPP      & 31.26                             & 38.39                              & 60.79                              & 69.05                             & 14.19                             \\ 
w/o MDE      & 26.39                             & 34.27                              & 55.58                              & 66.72                             & 12.75                             \\ 
w/o Position & 29.42                             & 35.29                              & 58.26                              & 68.18                             & 13.22                             \\ \hline
\end{tabular}}
\caption{Performance with Oracle concepts extracted from different methods, using BM25 as retriever.}
\label{ablation-ext}
\end{table}

\noindent \textbf{Ablation on legal concept generation models :} We compare different long-context models such as LongT5 \cite{guo2022longt5}, SLED-BART \cite{ivgi2023efficient}, and LED \cite{beltagy2020longformer}. We calculate the coverage of the generated concepts with respect to the reasoning section at both the word-level and concept (phrase) level in Tab. \ref{ablation-gen}. We notice LongT5 outperforming LED, which is semi-pretrained and initialized by repeatedly copying BART without undergoing end-to-end pre-training for longer sequences, thereby hindering its performance in long-context scenarios. Similarly, SLED employs a short-range pre-trained model like BART and uses a chunking approach on the encoder side to handle longer lengths, underscoring the necessity for long-context pre-training to effectively capture context for this task.

\begin{table}[]
\centering
\begin{tabular}{|l|c|c|}
\hline
       & \textbf{Word Cov.}   & \textbf{Concept Cov.} \\ \hline
LongT5    & 51.56    &  39.24    \\ 
SLED-BART & 46.12    &  34.46  \\ 
LED       & 44.29    &  30.29   \\ \hline
\end{tabular}
\caption{Coverage of generated concepts with respect to reasoning section for different generation models.}
\label{ablation-gen}
\end{table}

\subsection{Case Study}
\textbf{Example 1: The Case of O.P. v. THE REPUBLIC OF MOLDOVA\footnote{https://hudoc.echr.coe.int/?i=001-212690}}: Some of the key concepts extracted by our DPP framework are \textit{speediness of the requisite judicial controls, reasonableness of the suspicion,  speedy judicial decision, lawfulness of detention, reasonable will, deprivation of liberty, non‑pecuniary damage}. The concepts generated by LongT5 include \textit{right to liberty and security, lawful arrest or detention, reasonable suspicion, speediness of review}. These extracted and generated concepts certainly capture semantic intents which can  facilitate to retrieve relevant prior cases for this query, demonstrating the effectiveness of LeCoPCR.

\noindent \textbf{Example 2: The Case of BANCSÓK AND LÁSZLÓ MAGYAR v. HUNGARY\footnote{https://hudoc.echr.coe.int/?i=001-212669}} : The extracted key concepts are \textit{inhuman or degrading treatment or punishment, constitutional complaint, effective remedy, unprecedented delay,  potential effectiveness of the remedy, punitive element of punishment, de facto reducible , State’s margin of appreciation, general pardon procedure, apply for parole, prospect of release, progress towards rehabilitation, legitimate penological grounds, non-pecuniary damage}. While the generated ones are \textit{Life imprisonment, degrading punishment, periodic review, penological grounds, compassionate grounds, conditional release of the prisoner, medical assistance, rehabilitation}. These both concept lists reaffirm our hypothesis of LeCoPCR to boost retrieval performance.

\section{Conclusion}
We introduced LeCoPCR, a novel approach for PCR that leverages legal concepts to enhance the model's understanding of semantic intent that dictates relevance. We addressed the challenge of unavailability of annotated legal concepts by employing a weak supervision approach to extract key concepts from the reasoning section using DPP with masked-document-to-document matching and citation proximity, for quality assessment. Our results on the ECtHR-PCR dataset validate LeCoPCR's effectiveness. While our approach tackles relevance intent at a coarse-grained level for each query case, future research could explore fine-grained intent analysis for each cited case within a query case.

\section*{Limitations}
We assess LeCoPCR using the ECtHR-PCR dataset, which focuses on European Court of Human Rights judgments. While our approach may be generally applicable to other legal jurisdictions, its suitability across diverse legal systems, citation practices and semantic complexities warrants further investigation. Additionally, while LeCoPCR enhances retrieval explainability through generated legal concepts, our current evaluation, reliant on exact match-based coverage of the reasoning section, may not fully capture semantic nuances. Hence, a thorough human study is necessary to comprehensively gauge LeCoPCR's explainability.

A limitation of LeCoPCR is its exclusive focus on textual content, overlooking potential insights from the citation network's interconnected relationships, which offer a broader global perspective on case law. Future research should explore integrating citation structures to enhance retrieval performance. Another challenge lies in the temporal evolution of case law, reflecting changes in norms, societal attitudes, and the temporal nature of precedents, which may be overruled, thereby impacting the law's scope. Consequently, future approaches should account for the temporal aspects of precedents to model relevance effectively.

It's crucial to recognize that PCR datasets are constructed based on cases cited in the reasoning section of the query case, potentially leading to biases such as selective citation or the omission of relevant precedents. Thus, the effectiveness of LeCoPCR could be influenced by the dataset's representation, necessitating human evaluation to gauge the utility of retrieved precedents. However, conducting such evaluations poses challenges due to the lengthy and complex legal text, requiring annotators with deep expertise in ECHR jurisprudence, thus making it resource-intensive.

Furthermore, while the models explored in this study prioritize recall by acting as pre-fetchers to ensure retrieving all relevant cases, end-users typically expect a high-precision retrieval system that precisely identifies a smaller set of relevant cases. Achieving this requires an additional re-ranking step in the retrieval pipeline to optimize the precision of the ranked list. In this study, we focus on the initial retrieval step, leaving the development of a re-ranker component for future research.

\section*{Ethics Statement}
Our experiments utilize a dataset of ECHR decisions, which is publicly available and are sourced from the public court database HUDOC\footnote{\url{https://hudoc.echr.coe.int}},which includes real names and lacks anonymization. While there are concerns about the non-anonymized nature of this information, we do not foresee any significant harm beyond the disclosure. We believe that developing effective PCR systems is vital for supporting legal professionals in managing the growing caseload, underscoring the need for research to prioritize enhancing legal services and democratizing law \cite{tsarapatsanis2021ethical}. However, it is crucial to address various shortcomings and ensure the responsible and ethical deployment of legal-oriented technologies.

Furthermore, it is important to recognize that using historical data to train retrieval models may introduce biases, and leveraging pre-trained encoders can inherit biases encoded within them. Ensuring that PCR models intended for practical deployment do not inadvertently perpetuate or exacerbate existing biases in the legal system, such as racial or gender bias, is imperative. These models must undergo scrutiny against applicable equal treatment imperatives regarding their performance, behavior, and intended use, emphasizing the significance of fairness and accountability in the development and deployment of legal AI technologies.

\bibliography{custom}

\appendix
\section{Implementation Details}
\label{impl}
For legal concept generation model, we set the maximum input and output sequence length to 4096 and 256. We train with a learning rate of 1e-4  with a scheduler that warms up from zero during the first 10\% of the steps and then linearly decays back to zero for the remaining steps. The whole is model is trained for 10 epochs. We use  four beams and repetition penalty of 2.0. For BM25 retriever, we set the value of K1 and b to 1.2 and 0.75 respectively. For dense retriever, we use the maximum sequence length of 4096 and use 7 negatives per positive.  We use learning rate of 1e-4 and is trained for 3 epochs. We use FAISS \cite{johnson2019billion}, an open-source library for building efficient datastore construction for similarity search during dense retrieval. All models are optimized end-to-end using Adam \cite{kingma2014adam}.

\end{document}